\documentclass[runningheads]{llncs}
\usepackage{graphicx}

\usepackage{tikz}
\usepackage{comment}
\usepackage{amsmath,amssymb} 
\usepackage{color}
\usepackage{multirow}
\usepackage{wrapfig}

\usepackage[hyphens]{url}

\usepackage[font=small,skip=6pt]{caption}
\begin{document}
\pagestyle{headings}
\mainmatter
\def\ECCVSubNumber{2057}  

\title{AvatarCap: Animatable Avatar Conditioned Monocular Human Volumetric Capture} 

\titlerunning{AvatarCap} 
\authorrunning{Z. Li et al.} 
\author{Zhe Li \and Zerong Zheng \and Hongwen Zhang \and Chaonan Ji \and Yebin Liu}
\institute{Department of Automation, Tsinghua University, China}

%
%
%
\maketitle

\begin{abstract}
To address the ill-posed problem caused by partial observations in monocular human volumetric capture, we present AvatarCap, a novel framework that introduces animatable avatars into the capture pipeline for high-fidelity reconstruction in both visible and invisible regions.
Our method firstly creates an animatable avatar for the subject from a small number  ($\sim$20) of  3D scans as a prior. 
Then given a monocular RGB video of this subject, our method integrates information from both the image observation and the avatar prior, and accordingly reconstructs high-fidelity 3D textured models with dynamic details regardless of the visibility. 
To learn an effective avatar for volumetric capture from only few samples, we propose GeoTexAvatar, which leverages both geometry and texture supervisions to constrain the pose-dependent dynamics in a decomposed implicit manner. 
An avatar-conditioned volumetric capture method that involves a canonical normal fusion and a reconstruction network is further proposed to integrate both image observations and avatar dynamics for high-fidelity reconstruction in both observed and invisible regions.
Overall, our method enables monocular human volumetric capture with detailed and pose-dependent dynamics, and the experiments show that our method outperforms state of the art. 
\end{abstract}
\begin{figure}[t]
    \centering
    \includegraphics[width=1.0\linewidth]{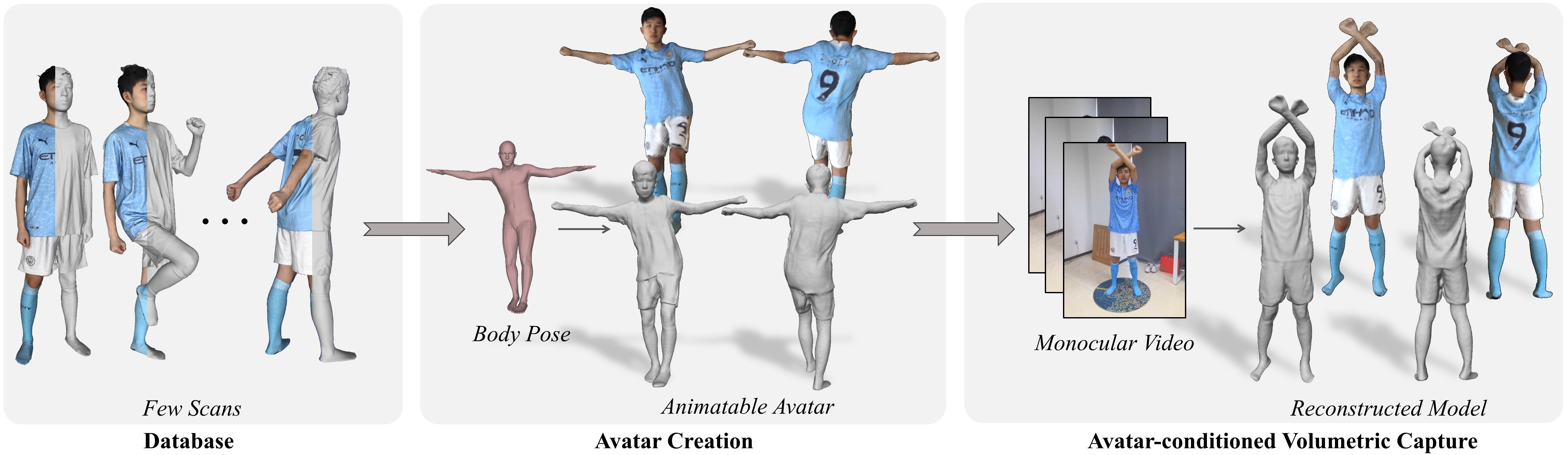}
    \caption{\textbf{Overview of AvatarCap.} We present AvatarCap that leverages an animatable avatar learned from only a small number ($\sim$20) of scans for monocular human volumetric capture to realize high-fidelity reconstruction regardless of the visibility.}
    \label{fig:teaser}
\end{figure}

\section{Introduction}

Human volumetric capture has been a popular research topic in computer vision for decades due to its potential value in Metaverse, holographic communication, video games, etc. Multi-view systems~\cite{bradley2008markerless,gall2009motion,liu2009point,brox2009combined,liu2011markerless,ye2012performance,mustafa2015general,dou2016fusion4d,pons2017clothcap,leroy2017multi,yu2021function4d,hong2021stereopifu,suo2021neuralhumanfvv,zheng2021deepmulticap,shao2021doublefield} can reconstruct high-resolution 3D human models using multiple RGB(D) sensors, but the sophisticated setup restricts their deployment in practice. 
To overcome this limitation, researchers have developed various technologies for monocular human reconstruction based on template tracking~\cite{zollhofer2014real,guo2015robust,habermann2019livecap,habermann2020deepcap}, volumetric fusion~\cite{newcombe2015dynamicfusion,yu2018doublefusion,su2020robustfusion} or single-image reconstruction~\cite{zheng2019deephuman,saito2019pifu,saito2020pifuhd,huang2020arch,li2020monocular,he2021arch++}. 

Despite the rapid development in monocular volumetric capture, most of the existing methods mainly focus on reconstructing visible surfaces according to direct observations and fail to recover the dynamic details in invisible regions. 
POSEFusion~\cite{li2021posefusion} addressed this limitation via integrating keyframes of similar poses from the whole RGBD sequence for invisible region reconstruction.   
However, it requires the subject to perform similar motions for multiple times facing different directions. 
What's worse, the fused invisible details are copied unaltered from other depth frames, thus suffering from poor pose generalization.

How to recover temporally coherent and pose-dependent details on invisible surfaces is an urgent and essential problem in monocular human voluemtric capture. Recently, many works on pose-driven human avatars have arisen in the community. They create animatable avatars from various inputs, including scans \cite{deng2020nasa,saito2021scanimate,ma2021scale,ma2021power,chen2021snarf}, multi-view RGB videos \cite{peng2021animatable,liu2021neural} and monocular depth measurements \cite{burov2021dynamic,wang2021metaavatar}. In this paper, our key insight is that the pose-driven dynamics of person-specific avatars are exactly what is missing in monocular human volumetric capture. With this in mind, we propose \textbf{\textit{AvatarCap}}, the first pipeline that combines person-specific animatable avatars with monocular human volumetric capture. 
Intuitively, the avatar encodes a data-driven prior about the pose-dependent dynamic details, which can compensate for the lack of complete observation in monocular inputs, enabling high-quality reconstruction of 3D models with dynamic details regardless of visibility.

Although introducing person-specific avatars into volumetric capture adds overhead in pipeline preparation, we believe that a data-driven prior of pose-dependent dynamics is indispensable for the future dynamic monocular human capture. In this paper, to make a trade-off between the ease of data acquisition and reconstruction quality, we choose to use only a small number ($\sim$ 20) of textured scans as the database.
Note that it is challenging to learn a generalized avatar from only few scans, 
and state-of-the-art methods typically require hundreds of scans for creating one avatar~\cite{saito2021scanimate,ma2021power}. 
If only twenty scans are available, their results tend to be overfit and lack geometric details because they condition all the surface details (including pose-dependent and pose-agnostic ones) on the pose input. To address this challenge, we propose \textbf{\textit{GeoTexAvatar}}, a decomposed representation that guarantees detail representation power and generalization capability.
To be more specific, our representation distills pose-agnostic details as much as possible into a common implicit template~\cite{zheng2021deep}, and models the remaining pose-driven dynamics with a pose-conditioned warping field. Such a disentanglement promotes better generalization since a large portion of geometric details are factored out as the common template and consequently the pose-dependent warping field is much easier to learn. 
On the other hand, previous methods rely on solely geometric cues to learn the conditional warping fields~\cite{zheng2021deep}, but we find that it is not enough because many types of cloth dynamics (e.g., cloth sliding) cannot be supervised by only geometry due to the ambiguity when establishing geometric correspondences. Therefore, we introduce an extra texture template represented by NeRF \cite{mildenhall2020nerf} to jointly constrain the pose-dependent warping field using both geometry and texture supervisions, which makes it possible to learn an accurate pose-conditioned warping field. As a result, the proposed GeoTexAvatar can not only preserve more details but also produce more reasonable pose-dependent dynamics for animation.

However, it is still not trivial to leverage the animatable avatar in the monocular capture pipeline. 
The main reason is the huge domain gap between the avatar prior and the monocular color input without any explicit 3D information. Fortunately, a 2D normal map with plentiful details can be extracted from the monocular color image~\cite{saito2020pifuhd}, and we can use it to bridge the 3D avatar and the 2D RGB input. However, directly optimizing the avatar geometry using extremely dense non-rigid deformation \cite{sumner2007embedded} to fit the 2D normal map is difficult, if not infeasible, 
because it is ill-posed to force the surface to be consistent with the normal map without explicit 3D correspondences. 
To overcome this challenge, we propose \textbf{\textit{Avatar-conditioned Volumetric Capture}} that splits the integration between the avatar and the normal maps into two steps, i.e., canonical normal fusion and model reconstruction. 
Specifically, the canonical normal fusion integrates the avatar normal and the image-observed one on the unified 2D canonical image plane. In this procedure, we formulate the fusion as an optimization on both the rotation grids and the normal maps to correct low-frequency normal orientation errors caused by inaccurate SMPL~\cite{loper2015smpl} fitting while maintaining high-frequency details.
After that, a reconstruction network pretrained on a large-scale 3D human dataset~\cite{yu2021function4d} is used as a strong prior for producing a high-fidelity 3D human with full-body details from the fused normal maps.

This paper proposes the following contributions:
1) AvatarCap, a new framework that introduces animatable avatars into the monocular human volumetric capture pipeline to achieve detailed and dynamic capture  regardless of the visibility (Sec.~\ref{sec:overview}).
2) GeoTexAvatar, a new decomposed avatar representation that contains a pose-agnostic Geo-Tex implicit template and a pose-dependent warping field to jointly constrain the pose-dependent dynamics using both geometry and texture supervisions for more detailed and well-generalized animation (Sec.~\ref{sec:avatar}).
3) Avatar-conditioned volumetric capture that contains a canonical normal fusion method and a reconstruction network to overcome the domain gap between the avatar prior and the monocular input for full-body high-fidelity reconstruction (Sec.~\ref{sec:capture}). Code is available at \textcolor{magenta}{\url{https://github.com/lizhe00/AvatarCap}}.

\section{Related Work}

\noindent\textbf{Template Tracking.} Given a monocular RGB(D) video, many works utilize a template to fit each frame using skeletal motion~\cite{magnenat1988joint} or non-rigid deformation~\cite{sumner2007embedded}. Specifically, \cite{li2009robust,zollhofer2014real,guo2015robust} solved the non-rigid warp field to track the input depth stream, while \cite{xu2018monoperfcap,guo2021human,zhi2020texmesh,he2021challencap} tracked the skeletal motion of the template to fit the monocular input.
LiveCap~\cite{habermann2019livecap} and DeepCap~\cite{habermann2020deepcap} jointly solved or inferred both skeletal and non-rigid motions from a monocular RGB video. 
MonoClothCap~\cite{xiang2020monoclothcap} built a statistical deformation model based on SMPL to capture visible cloth dynamics. However, these methods only focus on fitting the template to explain the image observation while neglecting the dynamics in invisible regions.

\noindent\textbf{Volumetric Fusion.} Meanwhile, to realize real-time reconstruction from a single depth sensor, Newcombe \textit{et al.}~\cite{newcombe2015dynamicfusion} pioneered to propose DynamicFusion that tracks and completes a canonical model in an incremental manner. It inspired a lot of following works~\cite{innmann2016volume,slavcheva2017killingfusion,guo2017real,yu2017BodyFusion,Slavcheva_2018_CVPR,chao2018ArticulatedFusion,yu2018doublefusion,Zheng2018HybridFusion,su2020robustfusion} to incorporate different body priors or other cues to improve the performance. However, similar to methods based on template tracking, these works do not take into account the dynamic deformations in invisible regions. SimulCap~\cite{yu2019simulcap} introduced cloth simulation into the volumetric fusion pipeline but its reconstruction quality is limited by a simple cloth simulator. POSEFusion~\cite{li2021posefusion} proposed to integrate multiple keyframes of similar poses to recover the dynamic details for the whole body, but this scheme leads to poor pose generalization, i.e., only those poses that are seen in different frames can be faithfully reconstructed. 

\noindent\textbf{Single-image Reconstruction.} 
Recently, researchers have paid more and more attention to recovering 3D humans from single RGB(D) images by volume regression~\cite{varol2018bodynet,jackson20183d,zheng2019deephuman}, visual hull~\cite{natsume2019siclope}, depth maps~\cite{gabeur2019moulding,smith2019facsimile}, template deformation~\cite{zhu2019detailed,alldieck2019tex2shape} and implicit functions~\cite{saito2019pifu,huang2020arch,li2020monocular,he2020geo,he2021arch++,xiu2021icon}. For the implicit function representation, PIFuHD~\cite{saito2020pifuhd} introduced normal estimation to produce detailed geometry. PaMIR~\cite{zheng2020pamir} and IPNet~\cite{bhatnagar2020combining} combined a parametric body model (e.g., SMPL~\cite{loper2015smpl}) into the implicit function to handle challenging poses. However, without direct observation, these methods only recover over-smoothed invisible geometry without details. NormalGAN~\cite{wang2020normalgan} inferred the back-view RGBD image from the input RGBD using a GAN~\cite{goodfellow2014generative} and then seamed them together. Unfortunately, the inferred details may be inconsistent with the pose or cloth type due to the limited variation in training data. 

\noindent\textbf{Animatable Human Avatar} To create animatable human avatars, previous methods usually reconstruct a template and then model the pose-dependent dynamics of the character by physical simulation~\cite{guan2012drape,stoll2010video} or deep learning~\cite{bagautdinov2021driving,xiang2021modeling,habermann2021real}. Recent works proposed to directly learn an animatable avatar from the database, including scans~\cite{ma2020learning,saito2021scanimate,ma2021scale,ma2021power,chen2021snarf}, multi-view RGB videos~\cite{liu2021neural,peng2021animatable} and depth frames~\cite{burov2021dynamic,wang2021metaavatar,dong2022pina}. 
These works usually require a large amount of data to train a person-specific avatar; when only a small number of scans are available, they suffer from overfitting and struggle with pose generalization. 
Wang \textit{et al.} ~\cite{wang2021metaavatar} learned a meta prior to overcome this issue, but it remains difficult to apply their method for texture modeling. 

\section{Overview}
\label{sec:overview}
\begin{figure}[t]
    \centering
    \includegraphics[width=\linewidth]{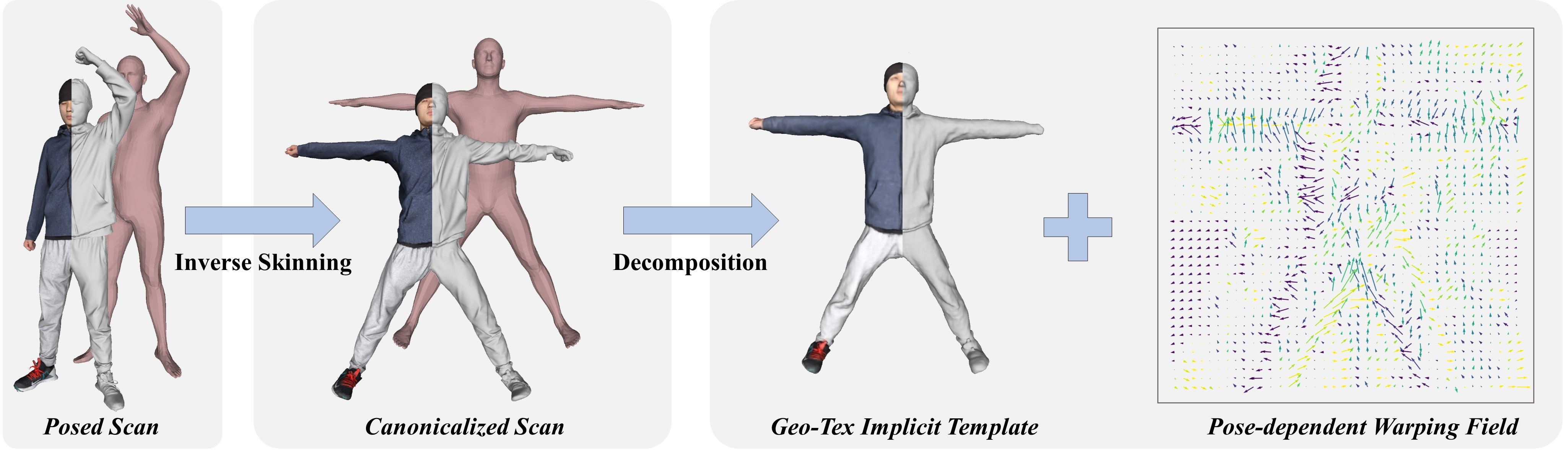}
    \caption{\textbf{Illustration of the GeoTexAvatar representation.} 
    We decompose the canonical scans into a pose-agnostic Geo-Tex implicit template and a pose-dependent warping field to enable joint supervisions by both geometry and texture for more detailed and well-generalized animation.
    }
    \label{fig:avatar representation}
\end{figure}

As shown in Fig.~\ref{fig:teaser}, the whole framework of AvatarCap contains two main steps:
\begin{enumerate}
    \item \textbf{Avatar Creation.} 
    Before performing monocular volumetric capture, we collect a small number ($\sim$ 20) of textured scans  for the subject as the database to construct his/her animatable avatar, which will be used to facilitate dynamic detail capture. 
    To create an avatar with realistic details and generalization capability, we propose GeoTexAvatar, a representation that decomposes the dynamic level set function~\cite{saito2021scanimate} into an implicit template (including occupancy~\cite{mescheder2019occupancy} and radiance~\cite{mildenhall2020nerf} fields) and a pose-dependent warping field, as shown in Fig.~\ref{fig:avatar representation}. We train the GeoTexAvatar network by supervising both the geometry and the texture using the textured scans.
    \item \textbf{Avatar-conditioned Volumetric Capture.}  
    With the avatar prior, we perform volumetric capture given the monocular RGB video input, as illustrated in Fig.~\ref{fig:recon pipeline}. 
    To address the domain gap between the avatar and the RGB input, we propose to use the surface normals as the intermediate to bridge each other.
    Specifically, we firstly estimate the visible normals from each RGB image, which are then mapped to the canonical space using the estimated SMPL pose \cite{kolotouros2019learning,zhang2021pymaf}. 
    Then we generate the canonical avatar with pose-dependent dynamics given the pose and render the canonical normal maps from both the front and back views. 
    The next step is to integrate the rendered normal maps with their image-based counterparts. 
    To do so, we propose canonical normal fusion, which aims to correct low-frequency local normal orientations while maintaining high-frequency details from image observations.
    Finally, a pretrained reconstruction network is used to produce a high-fidelity human model conditioned on the integrated normal maps.
\end{enumerate}
\section{Avatar Creation}
\label{sec:avatar}
In this section, our goal is to learn an animatable avatar for volumetric capture. 
Following the practice of  SCANimate~\cite{saito2021scanimate}, we fit SMPL to the raw 3D scans and transform them to a canonical pose via inverse skinning.  
We aim to construct an animatable avatar, represented as a pose-conditioned implicit function, from these canonicalized scans. 
Since only a small number ($\sim$ 20) of textured scans are available, we propose a decomposed implicit function to guarantee representation power and generalization capability (Sec.~\ref{sec:avatar:representation}), which allows us to better leverage the geometry and texture information of training data  (Sec.~\ref{sec:avatar:training}). 

\subsection{GeoTexAvatar Representation}
\label{sec:avatar:representation}

Our representation is built upon the pose-conditioned implicit function in SCANimate~\cite{saito2021scanimate}, which is defined as $f(\mathbf{x}_c,\boldsymbol{\theta})=s$,
where $s\in[0, 1]$ is an occupancy value, $\mathbf{x}_c$ is a 3D point in the canonical space and $\boldsymbol{\theta}$ is the SMPL pose parameters. The pose-dependent surface is represented by the level set of this implicit function:  $f(\mathbf{x}_c,\boldsymbol{\theta})=0.5$. However, such an entangled representation conditions all the surface dynamics, including the pose-dependent deformations and the pose-agnostic details, on the pose input. Consequently, the animation results tend to lack pose-agnostic details when given an unseen pose.

In order to not only model the pose-dependent deformations but also preserve the pose-agnostic details among different training samples, we propose a decomposed representation based on \cite{zheng2021deep}:
\begin{equation}
    T_\text{Geo}(W(\mathbf{x}_c,\boldsymbol{\theta}))=s,
    \label{eq:geometric dit}
\end{equation}
where $W(\mathbf{x}_c,\boldsymbol{\theta})=\mathbf{x}_c+{\rm\Delta} W(\mathbf{x},\boldsymbol{\theta})$ represents the pose-dependent warping field that takes the pose parameters and a point as input and returns its template position, and $T_\text{Geo}(\cdot)$ is the pose-agnostic occupancy template. 

Note that previous avatars learned from scans \cite{saito2021scanimate,ma2021scale,ma2021power} ignore the texture information even though their databases contain texture. However, we find the texture is essential to constrain the pose-dependent cloth deformations, because only geometrically closest constraints cannot establish correct correspondences, especially for common tangential cloth motions (e.g., cloth sliding). Therefore, we further introduce an extra texture template using the neural radiance field \cite{mildenhall2020nerf} (NeRF) in the same decomposed manner, i.e.,
\begin{equation}
    T_\text{Tex}(W(\mathbf{x}_c,\boldsymbol{\theta}))=(\sigma,\mathbf{c}),
    \label{eq:texture dit}
\end{equation}
where $T_\text{Tex}(\cdot)$ is a template radiance field that maps a template point to its density $\sigma$ and color $\mathbf{c}$. Note that we utilize the template NeRF to represent the scan texture without view-dependent variation, so we discard the view direction input. Thanks to the decomposition (Eq.~\ref{eq:geometric dit} \& Eq.~\ref{eq:texture dit}), our avatar representation, dubbed \textit{GeoTexAvatar}, is able to jointly constrain the pose-dependent warping field $W(\cdot)$ with the Geo-Tex implicit template field ($T_\text{Geo}(\cdot)$ \& $T_\text{Tex}(\cdot)$) under the joint supervision of geometry and texture of training scans. 
Fig.~\ref{fig:avatar representation} is an illustration of our representation. 

Compared with state-of-the-art scan-based avatar methods  \cite{saito2021scanimate,ma2021power}, our representation shows two main advantages as demonstrated in Fig.~\ref{fig:cmp avatar}. 1) The decomposed representation can preserve more pose-agnostic details for animation. 2) The joint supervision of geometry and texture enables more reasonable pose-dependent deformations. What's more, the decomposed representation allows us to finetune the texture template for high-quality rendering, which is also an advantage over other entangled methods as shown in Fig.~\ref{fig:eval decomposition}.

\subsection{GeoTexAvatar Training}
\label{sec:avatar:training}
The training loss for our GeoTexAvatar network contains a geometry loss, a texture loss and a regularization loss for the warping field, i.e.,
$\mathcal{L} = \lambda_\text{geo}\mathcal{L}_\text{geo} + \lambda_\text{tex}\mathcal{L}_\text{tex} + \lambda_\text{reg}\mathcal{L}_\text{reg}$, 
where $\lambda_\text{geo}$, $\lambda_\text{tex}$ and $\lambda_\text{reg}$ are the loss weights.

\noindent\textbf{Geometry Loss.} $\mathcal{L}_\text{geo}$ penalizes the difference between the inferred occupancy and the ground truth:
\begin{equation}
    \mathcal{L}_\text{geo} = \frac{1}{|\mathcal{P}|}\sum_{\mathbf{x}_p\in \mathcal{P}}\text{BCE}\left(s(\mathbf{x}_p), s^*(\mathbf{x}_p)\right),
\end{equation}
where $\mathcal{P}$ is the sampled point set, $s(\mathbf{x}_p)$ and $s^*(\mathbf{x}_p)$ are inferred and ground-truth occupancy, respectively, and $\text{BCE}(\cdot)$ measures the binary cross entropy.

\noindent\textbf{Texture Loss.} To jointly train the NeRF template, we render the textured scans to different views for the supervision. $\mathcal{L}_\text{tex}$ measures the error between the color rendered by the network and the real one:
\begin{equation}
    \mathcal{L}_\text{tex}=\frac{1}{|\mathcal{R}|}\sum_{\mathbf{r}\in\mathcal{R}}\left\|\hat{C}(\mathbf{r})-C^*(\mathbf{r})\right\|^2,
\end{equation}
where $\mathcal{R}$ is the set of ray samples in the image view frustum, $\hat{C}(\cdot)$ is the volume rendering function as in \cite{mildenhall2020nerf}, and $C^*(\mathbf{r})$ is the ground-truth color.

\noindent\textbf{Regularization Loss.} $\mathcal{L}_\text{reg}$ constrains the warped points by $W(\cdot)$ to be close with the input because the canonical pose-dependent dynamics are usually small:
\begin{equation}
    \mathcal{L}_\text{reg}=\frac{1}{|\mathcal{P}\mathop{\cup}\mathcal{P}_\mathcal{R}|}\sum_{\mathbf{x}_c\in\mathcal{P}\mathop{\cup}\mathcal{P}_\mathcal{R}}\left\|{\rm\Delta}W(\mathbf{x}_c,\boldsymbol{\theta})\right\|^2,
\end{equation}
where $\mathcal{P}_\mathcal{R}$ is the sampled points along each ray in $\mathcal{R}$ during volume rendering.

\begin{figure}[t]
    \centering
    \includegraphics[width=\linewidth]{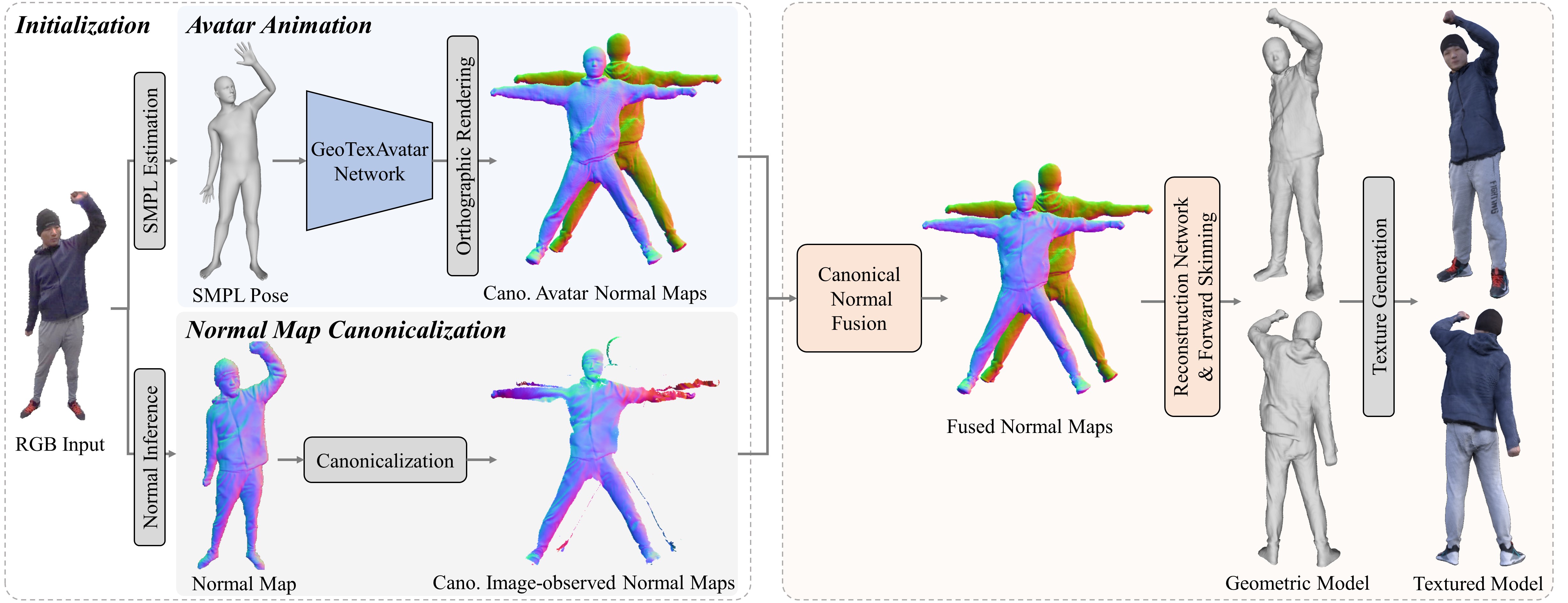}
    \caption{\textbf{Avatar-conditioned voluemtric capture pipeline.} Given a RGB image from the monocular video, we firstly infer the SMPL pose and normal map. Then the pose-driven GeoTexAvatar generates and renders canonical avatar normal maps, while the image-observed normal map is warped into the canonical space. The canonical normal fusion integrates both avatar and observed normals together and feeds the fused normal maps into the reconstruction network to output a high-fidelity 3D human model. Finally, a high-resolution texture is generated using the GeoTexAvatar network.}
    \label{fig:recon pipeline}
\end{figure}

\section{Avatar-conditioned Volumetric Capture}
\label{sec:capture}

Next, we move forward to the avatar-conditioned volumetric capture. The main difficulty lies in the enormous domain gap between the avatar representation and the input image, i.e., the image provides no 3D measurement to associate with the avatar geometry. As illustrated in Fig.~\ref{fig:recon pipeline}, to overcome this challenge, we propose to employ the normal maps as the intermediate representation to bridge the gap between the image inputs and the avatar prior. Specifically, we conduct the integration between the two modals on a unified canonical image plane, and then split the integration as canonical normal fusion and model reconstruction.

\textbf{Initialization.} Given a RGB image, our approach firstly prepares both the canonical avatar and image normal maps as illustrated in Fig.~\ref{fig:recon pipeline}. Specifically, 
1) Avatar Animation: The GeoTexAvatar network outputs the animated canonical avatar using the SMPL pose, then renders the front \& back canonical normal maps denoted as $\mathbf{F}_\text{avatar}$ \& $\mathbf{B}_\text{avatar}$, respectively.
2) Normal Map Canonicalization: In a parallel branch, the input RGB image is fed into a 2D convolutional network~\cite{wang2018high} to infer the normal map $\mathbf{N}$ that represents the visible details. Then it is mapped to the canonical space, with the results denoted as $\mathbf{F}_\text{image}$ and $\mathbf{B}_\text{image}$. Implementation details of the two steps can be found in the Supp. Mat.

\subsection{Canonical Normal Fusion}
\label{sec:capture:normal_optimization}
\begin{figure}[t]
    \centering
    \includegraphics[width=\linewidth]{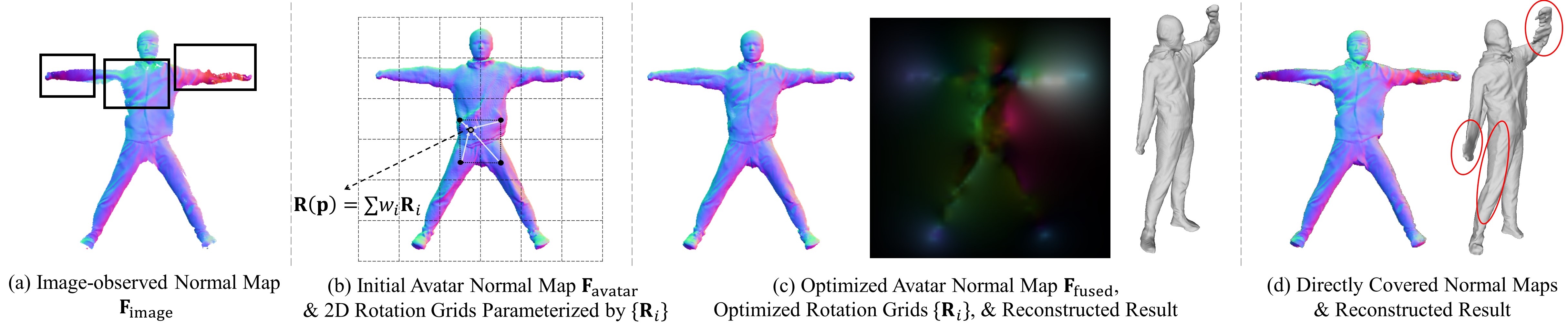}
    \caption{\textbf{Illustration of canonical normal fusion.} Directly replacing visible regions using image-observed normals causes severe reconstructed artifacts (d), while the proposed canonical normal fusion corrects the low-frequency local batch orientations and preserves high-frequency details for robust and high-fidelity reconstruction (c).}
    \label{fig:normal opt}
\end{figure}

Given the prepared avatar \& image-observed normal maps, we integrate them on the 2D canonical image plane.
However, directly replacing the avatar normals with the corresponding visible image-based ones is not feasible, because the canonicalized normal orientations may be incorrect due to the inaccurate SMPL estimation (e.g., rotation of the forearm) as shown in Fig.~\ref{fig:normal opt}(a), leading to severe artifacts in the reconstruction (Fig.~\ref{fig:normal opt}(d)). 
Therefore, we propose a new canonical normal fusion method to not only preserve high-frequency image-observed normals but also correct low-frequency local batch orientations.

Without loss of generality, we take the front avatar normal map and the image-based map  ($\mathbf{F}_\text{avatar}$ and $\mathbf{F}_\text{image}$) as the example to introduce the formulation.
As illustrated in Fig.~\ref{fig:normal opt}(a), $\mathbf{F}_\text{image}$ contains plentiful observed details estimated from the input color, but the orientations of normals are possibly incorrect due to SMPL estimation error. On the other hand, even though the visible region of $\mathbf{F}_\text{avatar}$ does not completely follow the image observation, the low-frequency normal orientations are accurate in the canonical space as shown in Fig.~\ref{fig:normal opt}(b).
To this end, we propose to optimize the avatar normal map $\mathbf{F}_\text{avatar}$ to integrate high-frequency details from the image-observed one $\mathbf{F}_\text{image}$ while maintaining its initial correct low-frequency orientations as shown in Fig.~\ref{fig:normal opt}(d).
To do so, we introduce 2D rotation grids to factor out the low-frequency orientation differences between $\mathbf{F}_\text{avatar}$ and $\mathbf{F}_\text{image}$, so that the remaining high-frequency details on $\mathbf{F}_\text{image}$ can be rotated back to $\mathbf{F}_\text{avatar}$ with correct orientations.
As illustrated in Fig.~\ref{fig:normal opt}(b), each grid is assigned a rotation matrix $\mathbf{R}_i\in SO(3)$, and the rotation of a 2D point $\mathbf{p}=(x,y)$ on the map is defined as $\mathbf{R}(\mathbf{p})=\sum_i w_i(\mathbf{p})\mathbf{R}_i$, a linear combination of $\{\mathbf{R}_i\}$ using bilinear interpolation, where $w_i(\mathbf{p})$ is the interpolation weight. With such a parameterization, we optimize the rotation grids $\{\mathbf{R}_i\}$ and avatar normal map $\mathbf{F}_\text{avatar}$ by minimizing
\begin{equation}
    E(\mathbf{R}_i,\mathbf{F}_\text{avatar})=\lambda_\text{fitting}E_\text{fitting}(\mathbf{R}_i,\mathbf{F}_\text{avatar}) + \lambda_\text{smooth}E_\text{smooth}(\mathbf{R}_i),
    \label{eq:normal opt formulation}
\end{equation}
where $E_\text{fitting}$ and $E_\text{smooth}$ are energies of misalignment between rotated avatar normals and observed ones and smooth regularization of grids, respectively.

\noindent\textbf{Fitting Term.} The fitting term measures the residuals between the avatar normal rotated by its transformation matrix and the target image-observed one:
\begin{equation}
    E_\text{fitting}(\mathbf{R}_i,\mathbf{F}_\text{avatar})=\sum_{\mathbf{p}\in\mathcal{D}}\left\|\mathbf{R}(\mathbf{p})\mathbf{F}_\text{avatar}(\mathbf{p})-\mathbf{F}_\text{image}(\mathbf{p})\right\|^2,
\end{equation}
where $\mathcal{D}$ is the valid intersection region of $\mathbf{F}_\text{avatar}$ and $\mathbf{F}_\text{image}$.

\noindent\textbf{Smooth Term}. The smooth term regularizes the rotation grids to be low-frequency by constrain the rotation similarity between adjacent grids:
\begin{equation}
    E_\text{smooth}(\mathbf{R}_i)=\sum_{i}\sum_{j\in\mathcal{N}(i)}\left\|\text{Rod}(\mathbf{R}_i)-\text{Rod}(\mathbf{R}_j)\right\|^2,
\end{equation}
where $\mathcal{N}(i)$ is the neighbors of the $i$-th grid, and $\text{Rod}: SO(3)\rightarrow so(3)$ maps the rotation matrix to the axis-angle vector.

\noindent\textbf{Delayed Optimization of} $\mathbf{F}_\text{avatar}$. We firstly initialize $\{\mathbf{R}_i\}$ as identity matrices. Note that both the avatar normal map $\mathbf{F}_\text{avatar}$ and rotation grids $\{\mathbf{R}_i\}$ are optimizable variables, so that the solutions are not unique. If we jointly optimizes both variables, $\mathbf{F}_\text{avatar}$ tends to be equal with $\mathbf{F}_\text{image}$ which is not desired. To this end, we firstly solve the low-frequency rotation grids $\{\mathbf{R}_i\}$, then optimize $\mathbf{F}_\text{avatar}$ to integrate high-frequency details from $\mathbf{F}_\text{image}$ with $\{\mathbf{R}_i\}$ fixed. As a result, we obtain the optimized $\mathbf{F}_\text{avatar}$ as the fused normal map $\mathbf{F}_\text{fused}$ with high-frequency details and correct low-frequency orientations as shown in Fig.~\ref{fig:normal opt}(c).

\subsection{Model Reconstruction}
\label{sec:capture:reconstruction}

\textbf{Geometric Reconstruction.} 
To reconstruct 3D geometry from the fused canonical normal maps $\mathbf{F}_\text{fused}$ \& $\mathbf{B}_\text{fused}$, we pretrain a reconstruction network on a large-scale 3D human dataset \cite{yu2021function4d}. With such a strong data prior, we can efficiently and robustly recover the 3D geometry with high-fidelity full body details from the complete normal maps.
The reconstruction network is formulated as an image-conditioned implicit function
$g(h(\pi(\mathbf{x});\mathbf{F}_\text{fused},\mathbf{B}_\text{fused}),\mathbf{x}_z)$,
where $\mathbf{x}$ is a 3D point in the canonical space, $h(\cdot)$ is a function to sample convoluted image features, $\pi(\cdot)$ is the orthographic projection, $\mathbf{x}_z$ is the z-axis value, and $g(\cdot)$ is an implicit function that maps the image feature and $\mathbf{x}_z$ to an occupancy value. We perform Marching Cubes \cite{lorensen1987marching} on this implicit function to reconstruct the canonical model, then deform it to the posed space by forward skinning.

\noindent\textbf{Texture Generation.}
Based on the GeoTexAvatar representation, we can generate the texture of the reconstructed geometry by mapping the radiance field to it. Specifically,
given a vertex $\mathbf{v}$ of the canonical model and its normal $\mathbf{n}_\mathbf{v}$, based on Eq.~\ref{eq:texture dit}, we can calculate its color using volume rendering in NeRF \cite{mildenhall2020nerf} with the camera ray $\mathbf{r}(t)=\mathbf{v}-t\mathbf{n}_\mathbf{v}$ and near and far bounds $-\delta$ and $\delta$ ($\delta>0$).

\begin{figure}[t]
    \centering
    \includegraphics[width=\linewidth]{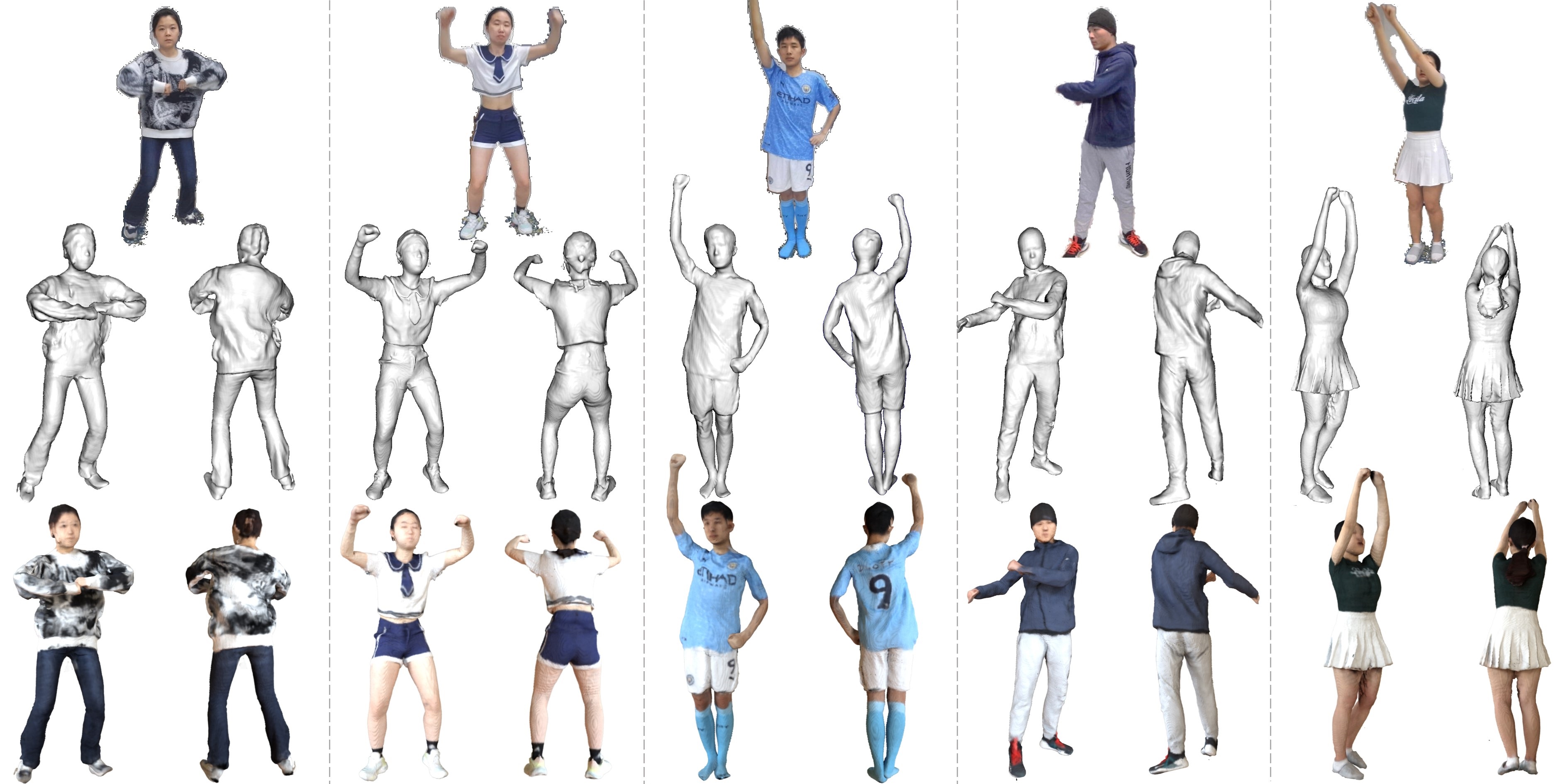}
    \caption{Example volumetric captured results of our method. From top to bottom are the monocular RGB input, geometric and textured results, respectively.}
    \label{fig:results}
\end{figure}

\section{Results}
The volumetric captured results of our method are demonstrated in Fig.~\ref{fig:results}. For the experiments, we collect textured scans of 10 subjects and their monocular videos, and partial scans are utilized as the evaluation dataset. More details about data preprocessing and implementation can be found in the Supp. Mat.

\subsection{Comparison}
\label{sec:comp}

\begin{figure}[t]
    \centering
    \includegraphics[width=\linewidth]{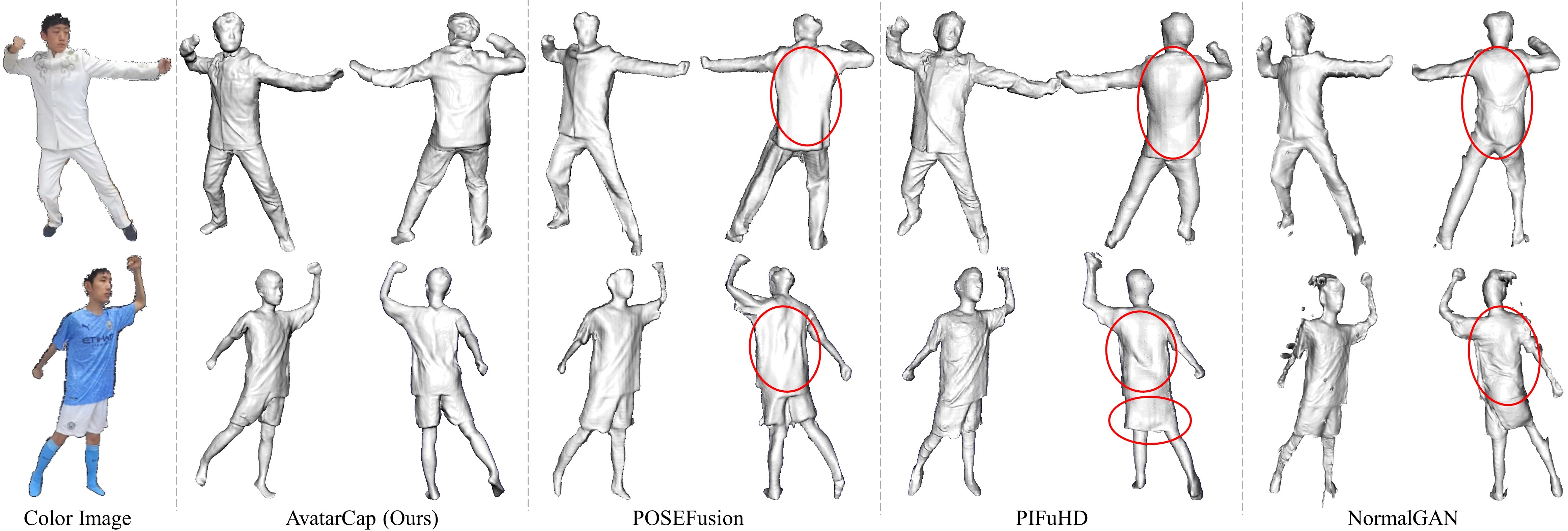}
    \caption{\textbf{Qualitative comparison against monocular volumetric capture methods.} We show reconstructed results of our method (AvatarCap), POSEFusion \cite{li2021posefusion}, PIFuHD \cite{saito2020pifuhd} and NormalGAN \cite{wang2020normalgan}. And our method outperforms others on the capture of pose-dependent dynamics in the invisible regions (red circles).}
    \label{fig:cmp capture}
\end{figure}

\noindent\textbf{Volumetric Capture.} As shown in Fig.~\ref{fig:cmp capture}, we compare AvatarCap, our whole volumetric capture framework, against state-of-the-art fusion and single-RGB(D)-image reconstruction methods, including POSEFusion \cite{li2021posefusion}, PIFuHD \cite{saito2020pifuhd} and NormalGAN \cite{wang2020normalgan}. We conduct this comparison on the sequences captured by one Kinect Azure to also compare against RGBD-based methods \cite{wang2020normalgan,li2021posefusion}, and all the learning-based methods are finetuned on the person-specific scans used in our avatar creation for fairness. Fig.~\ref{fig:cmp capture} shows that our method can achieve high-fidelity reconstruction with detailed observations and reasonable pose-dependent invisible dynamics. Though POSEFusion \cite{li2021posefusion} can integrate invisible surfaces from other frames, it entirely relies on each time captured sequence without pose generalization. PIFuHD \cite{saito2020pifuhd} only considers to recover visible details from the normal map inferred by the color input without pose-conditioned person-specific dynamics, so the invisible regions are usually oversmoothed. Though NormalGAN \cite{wang2020normalgan} can infer a plausible back RGBD map from the RGBD input, the inferred invisible appearance may be inconsistent with the person-specific dynamics.
We also conduct quantitative comparison on the testing dataset with ground-truth scans, and report the averaged errors in Tab.~\ref{tab:quant cmp capture}. 
Note that POSEFusion is a sequence-based method, but the testing scans are under discrete poses, so we only compare with the other methods.
Overall, our method achieves state-of-the-art capture on both quality and accuracy.

\begin{table}[t]
\centering
\begin{tabular}{l|ccc}
\hline
\multicolumn{1}{c|}{Metric/Method} & AvatarCap (ours) & PIFuHD \cite{saito2020pifuhd} & NormalGAN \cite{wang2020normalgan} \\ \hline
Chamfer Distance                   & \textbf{1.097}   & 3.400  & 2.852     \\
Scan-to-Mesh Distance              & \textbf{1.096}   & 3.092  & 2.855     \\ \hline
\end{tabular}
\caption{\textbf{Quantitative comparision of AvatarCap with PIFuHD \cite{saito2020pifuhd} and NormalGAN \cite{wang2020normalgan}}. We report the averaged Chamfer and Scan-to-Model distance errors ($\times 10^{-2}$ m) of differnt methods on the whole testing dataset.}
\label{tab:quant cmp capture}
\end{table}

\begin{figure}[t]
    \centering
    \includegraphics[width=\linewidth]{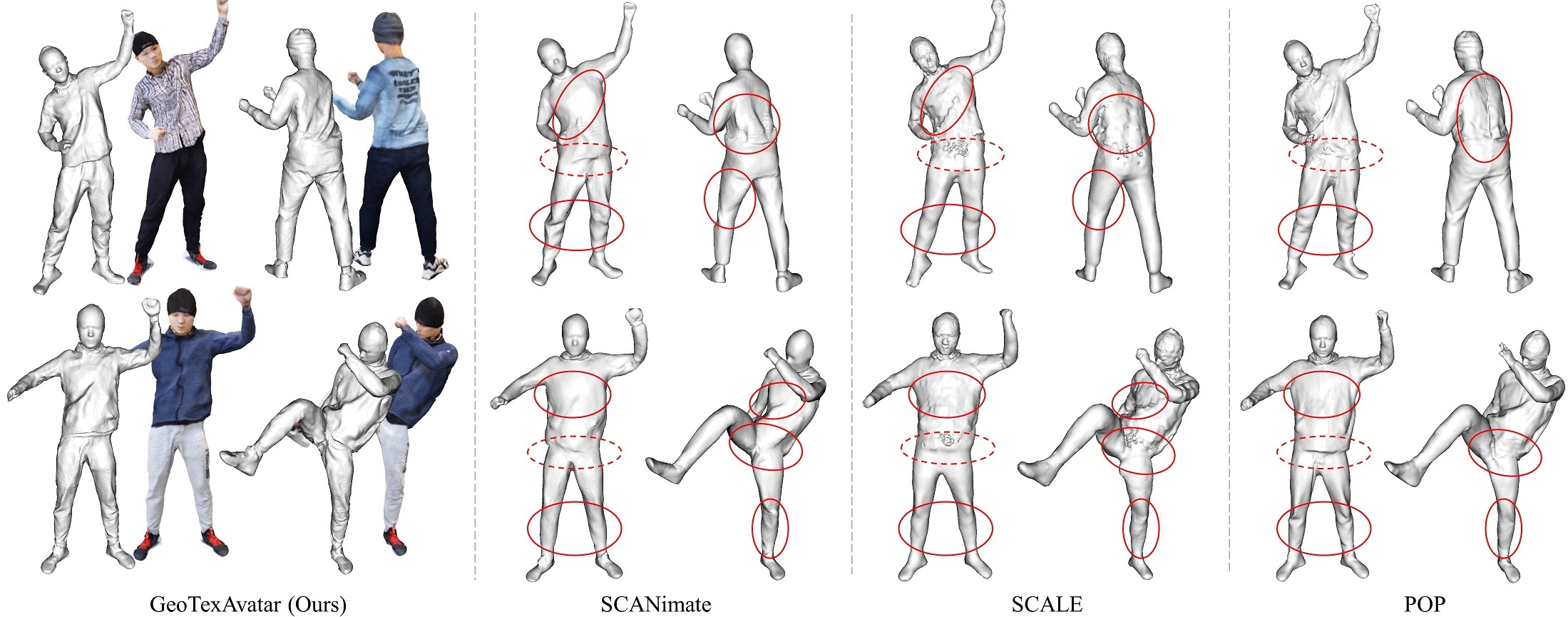}
    \caption{\textbf{Qualitative comparison against animatable avatar methods.} We show animated results of our method (also with high-quality texture), SCANimate \cite{saito2021scanimate}, SCALE \cite{ma2021scale} and POP \cite{ma2021power}. And our method shows the superiority on the modeling of wrinkles (solid circles) and pose-dependent cloth tangential motions (dotted circles).}
    \label{fig:cmp avatar}
\end{figure}

\begin{table}[t]

    \centering
    \begin{tabular}{c|cccc}
\hline
Case/Method           & GeoTexAvatar (Ours)  & SCANimate \cite{saito2021scanimate}            & SCALE \cite{ma2021scale}               & POP  \cite{ma2021power}                \\ \hline
HOODY\_1              & \textbf{6.29}        & 7.38                 & 8.19                 & 6.83                 \\
SHIRT\_1 & \textbf{2.80} & 5.72 & 4.72 & 3.08 \\ \hline
\end{tabular}
    \caption{\textbf{Quantitative comparison of GeoTexAvatar with SCANimate \cite{saito2021scanimate}, SCALE \cite{ma2021scale} and POP \cite{ma2021power}.} We report the averaged Chamfer distance errors ($\times 10^{-3}$ m) between the animated results of different methods and the ground-truth scans.}
    \label{tab:quant cmp avatar}
\end{table}

\noindent\textbf{Animatable Avatar.} As shown in Fig.~\ref{fig:cmp avatar}, we compare our avatar module, GeoTexAvatar, against state-of-the-art avatar works based on person-specific scans, SCANimate \cite{saito2021scanimate}, SCALE \cite{ma2021scale} and POP \cite{ma2021power}. Note that POP is a multi-subject-outfit representation, in this comparison we train it from scratch using the same few ($\sim$20) scans as other methods. Our method outperforms these methods on the recovery of dynamic details as well as tangential cloth motion benefiting from the proposed decomposed representation and joint supervisions of both geometry and texture, respectively. We further quantitatively evaluate the animation accuracy of GeoTexAvatar and other works on the testing dataset in Tab.~\ref{tab:quant cmp avatar}, and our method achieves more accurate animated results.

\begin{figure}[t]
    \centering
    \includegraphics[width=\linewidth]{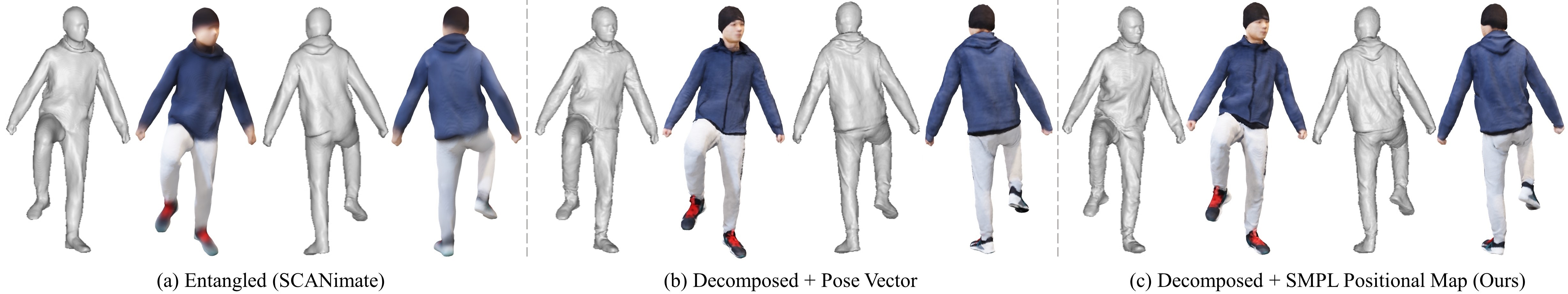}
    \caption{\textbf{Evaluation of the decomposed representation.} (a), (b) and (c) are the animated geometric and textured results of the entangled representation \cite{saito2021scanimate} and decomposed representations with pose-vector and positional-map encoding, respectively.}
    \label{fig:eval decomposition}
\end{figure}

\subsection{Evaluation}
\label{subsec:evaluation}

\noindent\textbf{Decomposed Representation of GeoTexAvatar.} 
We evaluate the proposed decomposed representation compared with the entangled one \cite{saito2021scanimate} in Fig.~\ref{fig:eval decomposition}.
Firstly, similar to SCANimate \cite{saito2021scanimate}, we choose the local pose vector as the pose encoding of the warping field in our representation. Compared with the entangled representation (Fig.~\ref{fig:eval decomposition} (a)), the decomposed one (Fig.~\ref{fig:eval decomposition} (b)) produces more detailed animation results, e.g., the zippers, facial and leg details, thanks to the decomposition of pose-dependent dynamics and pose-agnostic details. Besides, the decomposition allows us to finetune the texture template on a single scan to recovery high-quality texture, while the texture is totally blurred in the entangled learning. Furthermore, we empirically find that a SMPL positional map defined in the canonical space shows more powerful expression for pose-dependent dynamics than the local pose vector as shown in Fig.~\ref{fig:eval decomposition} (b) and (c).

\begin{figure}[t]
    \centering
    \includegraphics[width=\linewidth]{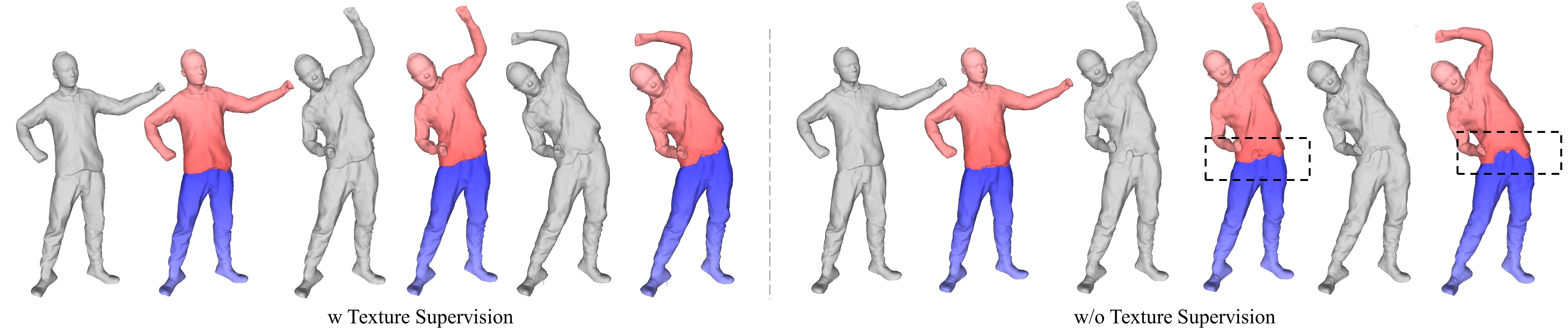}
    \caption{\textbf{Evaluation of the effectiveness of texture supervision in GeoTexAvatar.} We visualize the correspondences among different frames by the vertex color which indicates whether the vertex belongs to the upper or lower body.}
    \label{fig:eval tex super}
\end{figure}

\noindent\textbf{Texture Supervision in GeoTexAvatar.} We evaluate the effectiveness of texture supervision to the pose-dependent warping field by visualizing the correspondences during animation in Fig.~\ref{fig:eval tex super}. We firstly train the avatar network with and without texture template individually. To visualize the correspondences among animated results by different poses, we firstly generate the geometric template using \cite{lorensen1987marching} on $T_\text{Geo}(\cdot)$, then manually segment the template mesh as upper and lower body parts. Given a new pose, the avatar network outputs a canonical avatar model, then each vertex on this model can be warped to the template using the pose-dependent warping field. Finally, we determine whether the warped vertex belongs to the upper or lower part by its closest point on the template. Fig.~\ref{fig:eval tex super} demonstrates that the texture supervision can implicitly constrain the warping field by jointly learning an extra texture template $T_\text{Tex}(\cdot)$, thus enabling more reasonable pose generalization for animation. However, training only with geometry supervision results in overfitted animation due to the ambiguity when establishing correspondences under only the geometric closest constraint.

\begin{figure}[t]
    \centering
    \includegraphics[width=\linewidth]{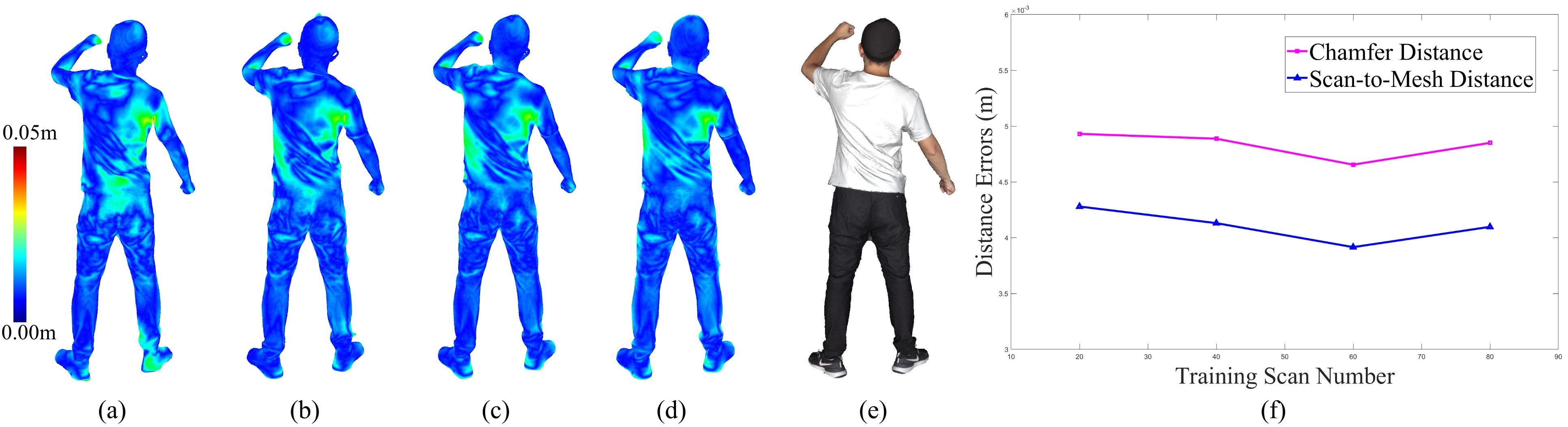}
    \caption{\textbf{Evaluation of the effect of the training scan number on the animation accuracy of the GeoTexAvatar.} From (a) to (d) are visualized vertex-to-surface error of animated results trained by 20, 40, 60 and 80 scans, respectively, (e) is the ground-truth scan, and (f) is the chart of the averaged Chamfer and Scan-to-Mesh distance errors on the whole testing dataset.}
    \label{fig:eval scan num}
\end{figure}

\noindent\textbf{Training Scan Number.} 
We quantitatively evaluate the effect of the training scan number on the animation accuracy of GeoTexAvatar. We choose one subset (``SHORT\_SLEEVE\_1'') of our dataset that contains 100 scans, and randomly choose 80 items as the training dataset, and the rest for evaluation. Fig.~\ref{fig:eval scan num} shows the visualized and numerical animation errors using different numbers of scans. More training samples does not always lead to more accurate results because the mapping from poses to cloth details may be one-to-many in the training dataset.

\begin{figure}[t]
    \centering
    \includegraphics[width=\linewidth]{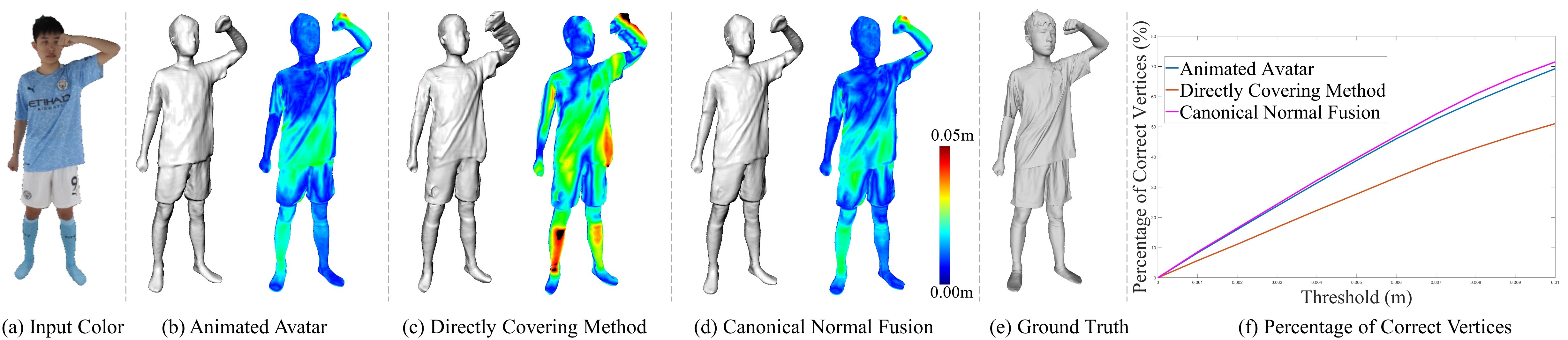}
    \caption{\textbf{Evaluation of canonical normal fusion.} We visualize the per-vertex point-to-surface errors between the reconstructed models and the ground truth, (f) is the percentage of correct vertices under different thresholds} 
    \label{fig:eval normal opt}
\end{figure}

\noindent\textbf{Canonical Normal Fusion.} We evaluate the proposed canonical normal fusion compared with the directly covering method both qualitatively and quantitatively. Fig.~\ref{fig:eval normal opt} (c) and (d) show the reconstructed results using directly covering and canonical normal fusion, respectively, as well as their per-vertex point-to-surface errors to the ground-truth scans. Due to the inaccurate SMPL estimation and the camera view difference with the orthographic hypothesis in normal map inference, the canonicalized normal tends to be fallacious. The directly covering method maintains the wrong image-observed normal, thus leading to inaccurate reconstruction and ghosting artifacts as shown in Fig.~\ref{fig:eval normal opt}(c). By contrary, the canonical normal fusion not only corrects the low-frequency orientations of canonicalized image normal, but also maintains the high-frequency details from the image observation, thus enabling the following accurate and high-fidelity reconstruction as shown in Fig.~\ref{fig:eval normal opt} (d) and (f).
\section{Discussion}
\textbf{Conclusion.} We present AvatarCap, a novel monocular human volumetric capture framework, that leverages an animatable avatar learned from only few scans to capture body dynamics regardless of the visibility. Based on the proposed GeoTexAvatar and avatar-conditioned volumetric capture, our method effectively integrates the information from image observations and the avatar prior. Overall, our method outperforms other state-of-the-art capture approaches, and we believe that the avatar-conditioned volumetric capture will make progress towards dynamic and realistic 3D human with the advance of animatable avatars.

\noindent\textbf{Limitation.} The main limitation of our method is the 3D scan collection, a possible solution is to capture scans using 3D self-portrait methods \cite{li20133d,li2020robust} with an RGBD camera. Moreover, our method may fail for loose clothes, e.g., long skirts, because the SMPL skeletons cannot correctly deform such garments.

\noindent\textbf{Acknowledgement.} This paper is supported by National Key R\&D Program of China (2021ZD0113501) and the NSFC project No.62125107. 

\appendix
In this supplementary material, we provide the implementation details of our method, more experiments and analysis. Please visit the project website\footnote{\textcolor{magenta}{\url{http://www.liuyebin.com/avatarcap/avatarcap.html}}} for more visualization of our results. 

\section{Implementation Details}
\subsection{Data Collection and Preprocessing}
\label{supp:data processing}
\begin{figure}[t]
    \centering
    \includegraphics[width=\linewidth]{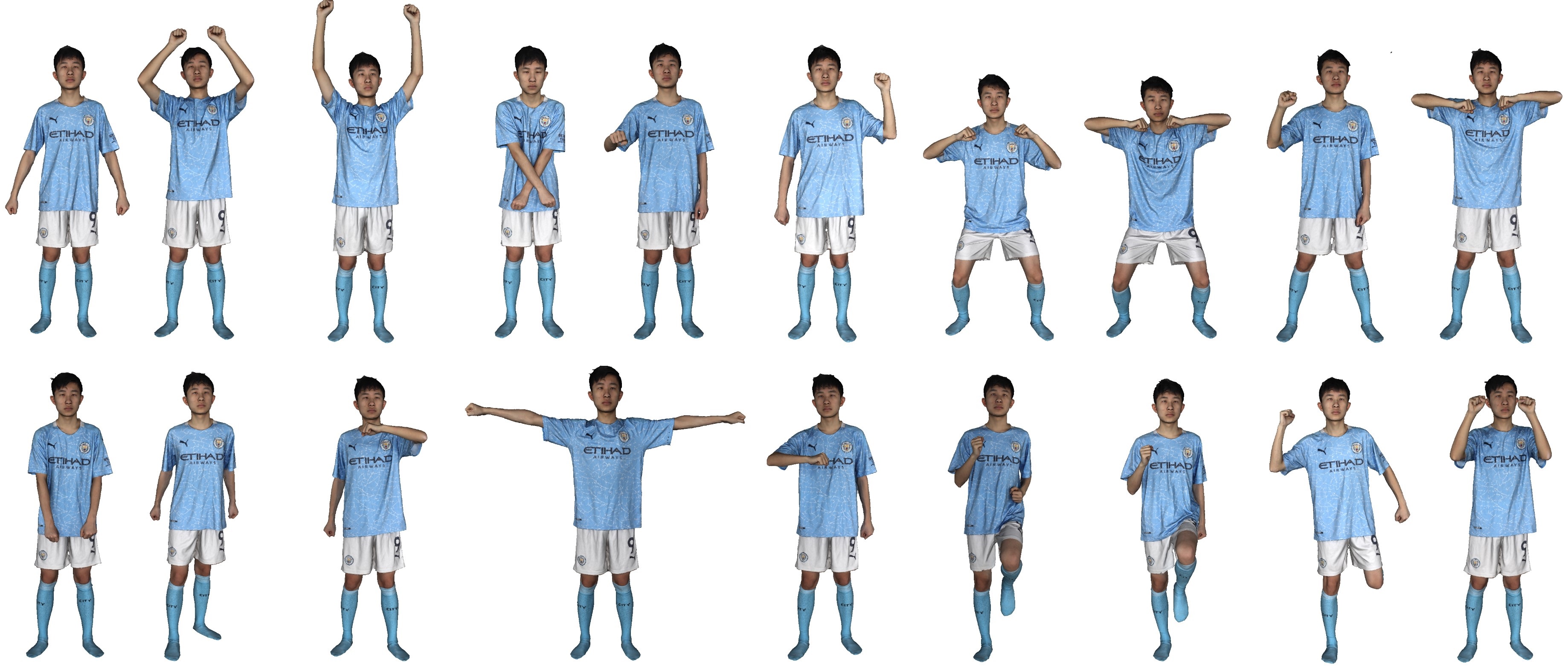}
    \caption{\textbf{Training scans of one subject.}}
    \label{fig:training scans}
\end{figure}

The textured scans are captured using a dense DLSR rig as the training database for creating the avatar as shown in Fig.~\ref{fig:training scans}. We firstly fit SMPL \cite{loper2015smpl} to each scan using \cite{EasyMocap}. Then the scan is deformed to the canonical pose following ARCH \cite{huang2020arch}. Different from directly learning SDF from the non-watertight canonicalized scans \cite{saito2021scanimate}, we non-rigidly deform a canonical SMPL to align with the scan for filling the holes, then utilize Poisson reconstruction \cite{kazhdan2006poisson} to generate watertight scans. Finally, to jointly train the texture template represented by NeRF \cite{mildenhall2020nerf}, we render the original textured scans from 60 views distributed uniformly in a circle.

Note that the texture and occupancy supervisions are not in the same space, i.e., the former is in the posed space while the later is in the canonical space. The reason for that is there may exist body part intersections on the original captured scans, e.g., the armpits, if we sample points around these regions in the posed space, the corresponding ground-truth occupancy values will be incorrect.

\subsection{GeoTexAvatar}
\label{supp:avatar}
\noindent\textbf{Network Architecture.}
The GeoTexAvatar network contains two modules, i.e., the Geo-Tex implicit template and the pose-conditioned warping field. The Geo-Tex implicit template is represented as an MLP, which takes a 3D template point with 10th-order positional encoding \cite{mildenhall2020nerf,tancik2020fourier} as input, and returns its occupancy, color and density. The template network consists of a shared MLP with (63, 256, 256, 256, 256, 256, 256, 256) neurons, a geometry MLP with (256, 128, 2) neurons and a color MLP with (256, 256, 128, 3) neurons. The geometry MLP jointly outputs the occupancy and density value; such an implicit representation is inspired by \cite{shao2021doublefield}. The last non-linear activation functions of occupancy, density and color MLPs are Sigmoid, ReLU and Sigmoid, respectively.

The pose-conditioned warping field consists of a positional map encoder $E(\cdot)$ and an offset decoder $D(\cdot)$:
\begin{equation}
    \Delta W(\mathbf{x}_c,\boldsymbol{\theta})=D(\mathbf{x}_c, B(\pi(\mathbf{x}_c); E(\mathbf{P}(\boldsymbol{\theta})))),
\end{equation}
where $\mathbf{x}_c$ is a canonical 3D point, $\mathbf{P}(\boldsymbol{\theta})$ is the rendered canonical SMPL positional map where the pixel value is the posed SMPL vertex position, $B(\cdot)$ is a bilinear sampling function to sample feature on the feature map $E(\mathbf{P}(\boldsymbol{\theta}))$ for $\mathbf{x}_c$, and $\pi(\cdot)$ is the orthographic projection to project $\mathbf{x}_c$ onto the 2D plane of the rendered positional map. 
To generate the positional map, we render the canonical SMPL from front and back views to generate two pixel-aligned positional maps,  then concatenate them together, and finally feed them to the positional map encoder $E(\cdot)$ followed by the offset decoder $D(\cdot)$. Following \cite{ma2021power}, the positional map encoder $E(\cdot)$ is a UNet \cite{ronneberger2015u} that contains seven [Conv2d, BatchNorm, LeakyReLU(0.2)] blocks, followed by seven [ReLU, ConvTranspose2d, BatchNorm] blocks, and it returns a $256\times256\times64$ feature map. The offset decoder is a MLP that takes the canonical point and corresponding feature as input, and it contains (3+64, 256, 256, 256, 256, 256, 256, 256, 3) neurons at each layer, respectively.
Note that in \cite{ma2021power} the SMPL positional map in defined in the SMPL UV space. We do not follow the practice in \cite{ma2021power} 
because we need to query the feature for the whole 3D space. Our definition also avoids the discontinuity in the UV space which causes the seam artifacts on the back of animated models in \cite{ma2021power}. 

\noindent\textbf{Training.}
We train the whole network in an end-to-end manner using the Adam \cite{adam} optimizer with a batch size of 4 for 30 epochs on $\sim$ 20 scans. The loss weights are set as $\lambda_\text{geo}=0.5,\lambda_\text{tex}=1.0,\lambda_\text{reg}=0.1$. The initial learning rates of the Geo-Tex implicit template and warping field are $1\times10^{-3}$ and $1\times10^{-4}$, respectively, and drop half every 20000 iterations. We initialize the warping field to output zero offsets, and at the first two epochs, we fix the warping field and only optimize the template network to obtain a coarse template. The training of one subject for creating an animatable avatar takes about two hours.

\subsection{Avatar-conditioned Volumetric Capture}
As shown in Fig.~3 of the main paper, the initialization of the volumetric capture contains avatar animation and normal map canonicalization.

\noindent\textbf{Avatar Animation.} Firstly, we can estimate SMPL pose from the monocular color input using SPIN \cite{kolotouros2019learning} or PyMAF \cite{zhang2021pymaf}. With the SMPL pose, we can generate canonical SMPL positional map as described in Sec.~\ref{supp:avatar}. We allocate a canonical volume that contains the canonical SMPL body. For each voxel, we feed its position and projected feature on the convoluted feature map to the network to evaluate its occupancy, then we perform Marching Cubes \cite{lorensen1987marching} on this occupancy volume to acquire a canonical geometric model. Finally, we render it from front and back views by orthographic projection to obtain front and back avatar normal maps.

\noindent\textbf{Normal Map Canonicalization.} In this branch, we firstly estimate the normal map from the monocular color input using pix2pixHD \cite{wang2018high} following PIFuHD \cite{saito2020pifuhd}. Then we deform the canonical avatar model using the estimated SMPL pose to the image/posed space, then project it onto the normal map to fetch a normal vector for each visible vertex. Similar to avatar animation, we render the fetched normals using the canonical avatar from the same front and back views by orthographic projection to obtain front and back image-observed normal maps.

With the above two steps in the initialization, we bridge the avatar and image information on the unified 2D canonical image plane.

\noindent\textbf{Canonical Normal Fusion} As introduced in Sec.~5.1 of the main paper, we formulate the fusion as an optimization, and in the energy function Eq.~6, we set $\lambda_\text{fitting}=1.0$ and $\lambda_\text{smooth}=1.0$, and we optimize it using Gauss-Newton algorithm for 50 iterations. The resolution of all the normal maps are $512\times512$, and the resolution of the rotation grids is $64\times64$.

\noindent\textbf{Model Reconstruction} We introduce a reconstruction network pretrained on a large-scale human dataset (THuman 2.0 \cite{yu2021function4d}) to leverage the data prior to infer the 3D model from the fused normal maps. Because the normal maps are in the canonical space, similar to Sec.~\ref{supp:data processing}, we deform all the original scans to the canonical pose by the SMPL registration. Then we render the canonicalized scan from front and back views to obtain normal maps. We sample 3D points randomly near surfaces and in the canonical volume as in PIFu \cite{saito2019pifu}, then calculate their occupancy values. With the rendered normal maps and sampled points, we train this network using the Adam \cite{adam} optimizer with a batch size of $4$ and a learning rate of $1\times10^{-3}$ for 240 epochs. The training takes about two days on one RTX 3090 GPU.

\subsection{Runtime Performance}
Given $\sim$20 textured scans of one subject, it takes about 0.5 hours for the data preprocessing and 2.0 hours for the avatar training. In the volumetric capture, the avatar animation, normal map canonicalization, canonical normal fusion, model reconstruction and texture generation cost about 1.0, 0.5, 1.2, 0.8, 3.0 secs, respectively. Overall, our method takes about 6 $\sim$ 7 secs for reconstructing one frame.

\begin{figure}[h]
    \centering
    \includegraphics[width=\linewidth]{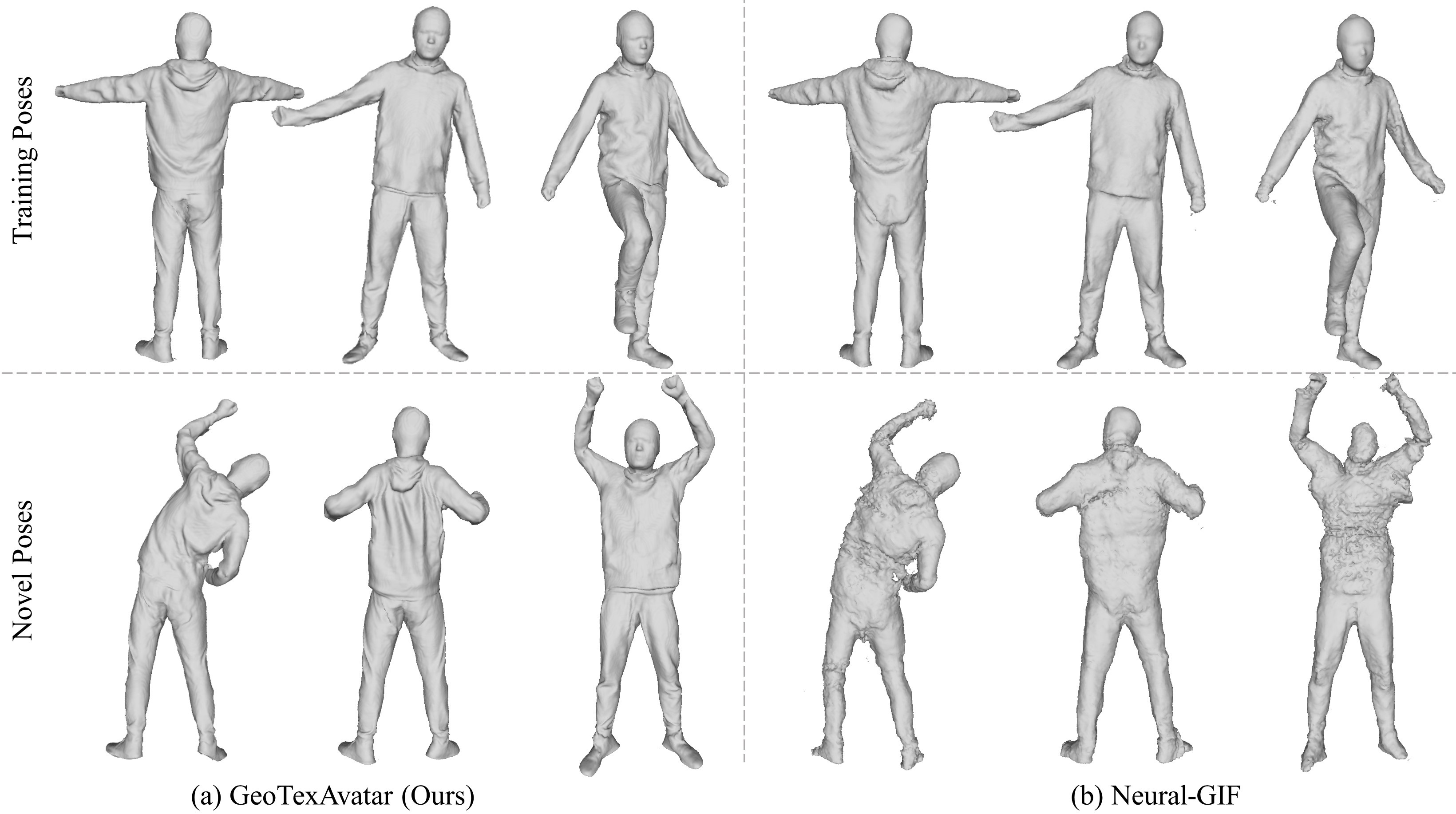}
    \caption{\textbf{Comparison between GeoTexAvatar and Neural-GIF \cite{tiwari2021neural}.} We show animated results by our GeoTexAvatar and Neural-GIF on both training and novel poses, respectively.}
    \label{fig:cmp neural_gif}
\end{figure}

\section{Additional Experiment}
\noindent\textbf{Comparison against Neural-GIF \cite{tiwari2021neural}.} We further compare our animatable avatar module, GeoTexAvatar, against another state-of-the-art scan-based avatar method, Neural-GIF \cite{tiwari2021neural}. Fig.~\ref{fig:cmp neural_gif} shows the animated results of our method and Neural-GIF on both training and novel poses, respectively. It shows that Neural-GIF suffers from overfitting, and cannot generalize the avatar trained on 22 scans to the novel poses. We hypothesize that the reasons include: 1) It is hard for the inverse skinning network in Neural-GIF to learn good generalization from only few examples, because its input coordinate is in the posed space where the skinning weight of the same position varies significantly when the SMPL pose changes; 2) Neural-GIF does not decompose the pose-agnostic details and the pose-dependent ones and it conditions both the displacement and canonical SDF networks on the pose input, thus all the surface details are driven by the pose input. Benefiting from the decomposition between pose-agnostic and pose-dependent details, our method realizes more robust and plausible pose generalization.

\begin{wrapfigure}[7]{r}{0.4\linewidth} 
\centering
\includegraphics[width=0.9\linewidth]{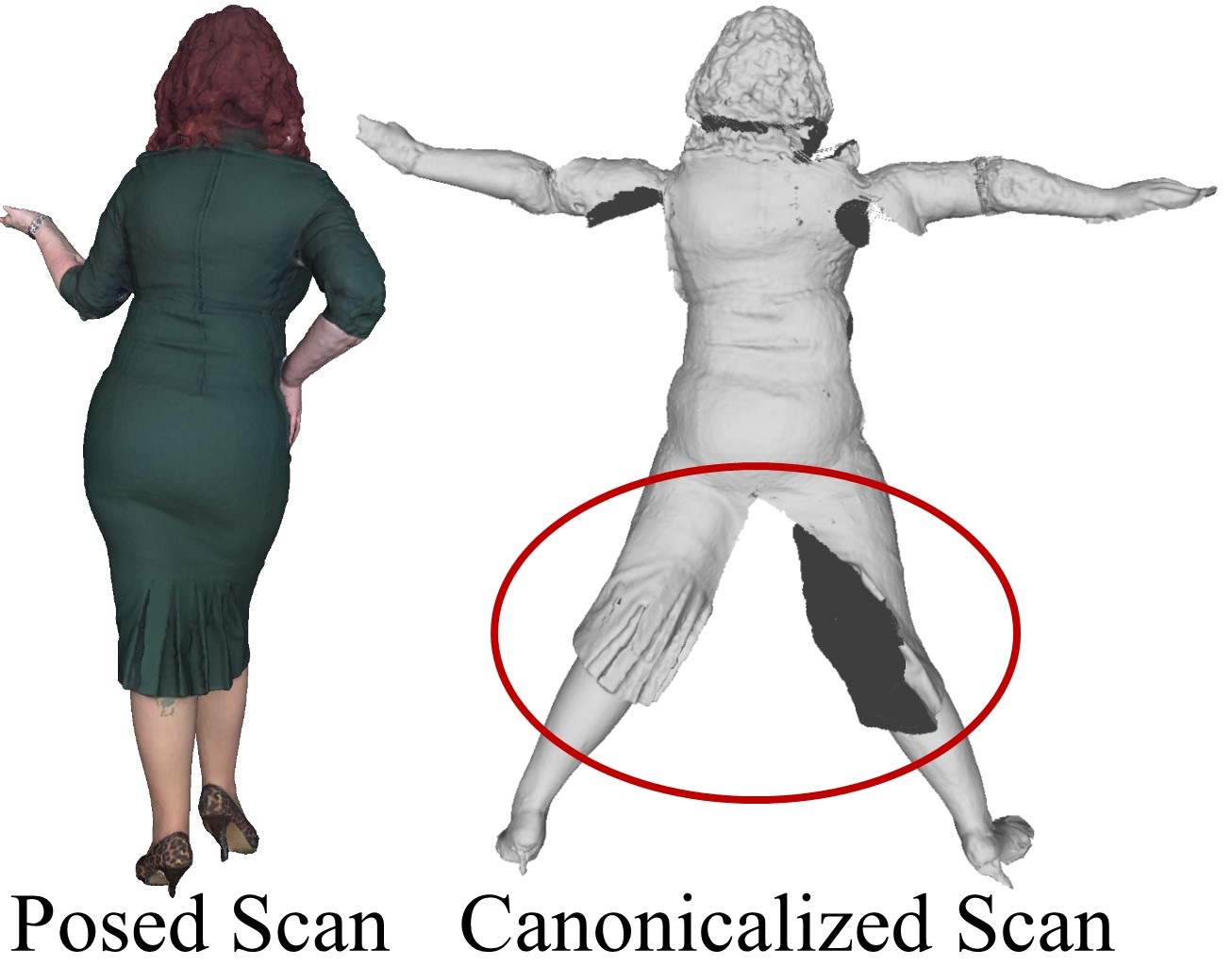}
\end{wrapfigure}

\noindent\textbf{Failure Case.}
Our method cannot handle loose clothes, e.g., long dresses, because the canonicalization step may fail for such garments as shown on the right. 
It remains difficult to deform the long dress to the canonical space using only SMPL skeletons, a possible solution is to parameterize the deformation by non-rigid embedded node graph \cite{sumner2007embedded}.

\clearpage
%
%
\bibliographystyle{splncs04}
\bibliography{egbib}
\end{document}